\documentclass[10pt,twocolumn,letterpaper]{article}

\usepackage[pagenumbers]{cvpr} % To force page numbers, e.g. for an arXiv version

\renewcommand{\paragraph}[1]{\vspace{.1em}\noindent\textbf{#1.}}

\definecolor{cvprblue}{rgb}{0.21,0.49,0.74}
\usepackage[pagebackref,breaklinks,colorlinks,allcolors=cvprblue]{hyperref}
\usepackage{booktabs,pifont}
\newcommand{\cmark}{\ding{51}} % ✓
\usepackage{enumitem}
\usepackage{listings}
\usepackage[T1]{fontenc}
\usepackage[scaled=.8]{beramono}   % Scale texttt. Default is .9
\usepackage{soul}
\usepackage{xspace}
\usepackage{amsmath}
\sethlcolor{white}
\usepackage{mathtools}
\usepackage{multirow}

\newcommand{\METHODNAME}{{\fontfamily{txtt}\selectfont {LCDrive}}\xspace}

\def\paperID{11814} % *** Enter the Paper ID here
\def\confName{CVPR}
\def\confYear{2026}

\title{Latent Chain-of-Thought World Modeling for End-to-End Autonomous Driving}

\author{
Shuhan Tan$^{1,2}$\thanks{Work done during an internship at NVIDIA.} \quad
Kashyap Chitta$^{2}$ \quad
Yuxiao Chen$^{2}$ \quad
Ran Tian$^{2}$ \quad
Yurong You$^{2}$ \quad
Yan Wang$^{2}$ \quad\\
Wenjie Luo$^{2}$ \quad
Yulong Cao$^{2}$ \quad
Philipp Kr\"ahenb\"uhl$^{1}$ \quad
Marco Pavone$^{2,3}$ \quad
Boris Ivanovic$^{2}$\\[0.5em]
$^{1}$UT Austin \quad
$^{2}$NVIDIA \quad
$^{3}$Stanford University\\[0.25em]
}

\begin{document}
\maketitle

\begin{abstract}
Recent Vision-Language-Action (VLA) models for autonomous driving explore inference-time reasoning as a way to improve driving performance and safety in challenging scenarios.
Most prior work uses natural language to express chain-of-thought (CoT) reasoning before producing driving actions.
However, text may not be the most efficient representation for reasoning. In this work, we present \textbf{L}atent-\textbf{C}oT-\textbf{Drive} (\METHODNAME): a model that expresses CoT in a \emph{latent} language that captures possible outcomes of the driving actions being considered.
Our approach unifies CoT reasoning and decision making by representing \emph{both} in an action-aligned latent space.
Instead of natural language, the model reasons by interleaving (1) action-proposal tokens, which use the same vocabulary as the model's output actions; and (2) world model tokens, which are grounded in a learned latent world model and express future outcomes of these actions.
We cold start latent CoT by supervising the model's action proposals and world model tokens based on ground-truth future rollouts of the scene.
We then post-train with closed-loop reinforcement learning to strengthen reasoning capabilities.
On a large-scale end-to-end driving benchmark, \METHODNAME achieves faster inference, better trajectory quality, and larger improvements from interactive reinforcement learning compared to both non-reasoning and text-reasoning baselines.
\end{abstract}

\section{Introduction}
\label{sec:intro}

End-to-end (E2E) autonomous driving aims to map raw, multi-view camera streams together with ego state, history, and high-level navigation commands \textit{directly} to future trajectories and low-level controls using a single policy~\cite{hu2023uniad,weng2024paradrive}.
A growing trend is to instantiate this policy as a \textit{Vision--Language--Action (VLA)} foundation model~\cite{kawaharazuka2025vlasurvey}, pre-trained on large-scale vision-language data and fine-tuned on driving logs.
Building on this trend, recent studies introduce \textit{inference-time reasoning} by generating a text-based chain-of-thought (CoT) before committing to actions~\cite{tian2024tokenize, wang2024drivecot, hwang2024emma, zhou2025autovla,wang2025alpamayo}.
While this is a natural choice following recent works on reasoning LLMs~\cite{wei2022chain}, a textual CoT presents several limitations when applied to driving.
First, natural language is ill-suited for representing spatiotemporal geometry and multi-agent interactions, which are central to driving decision-making.
Second, autoregressively generating long chains of text introduces nontrivial latency, making real-time deployment challenging.
Furthermore, the generated actions may significantly diverge from the preceding language rationales (e.g., the text states ``go left'' yet the action indicates a right turn) due to weak action--text alignment~\cite{wang2025alpamayo}.
Accordingly, we argue that text is not the most suitable substrate in driving VLA models.

\begin{figure}[t]
    \centering
    \includegraphics[width=\linewidth]{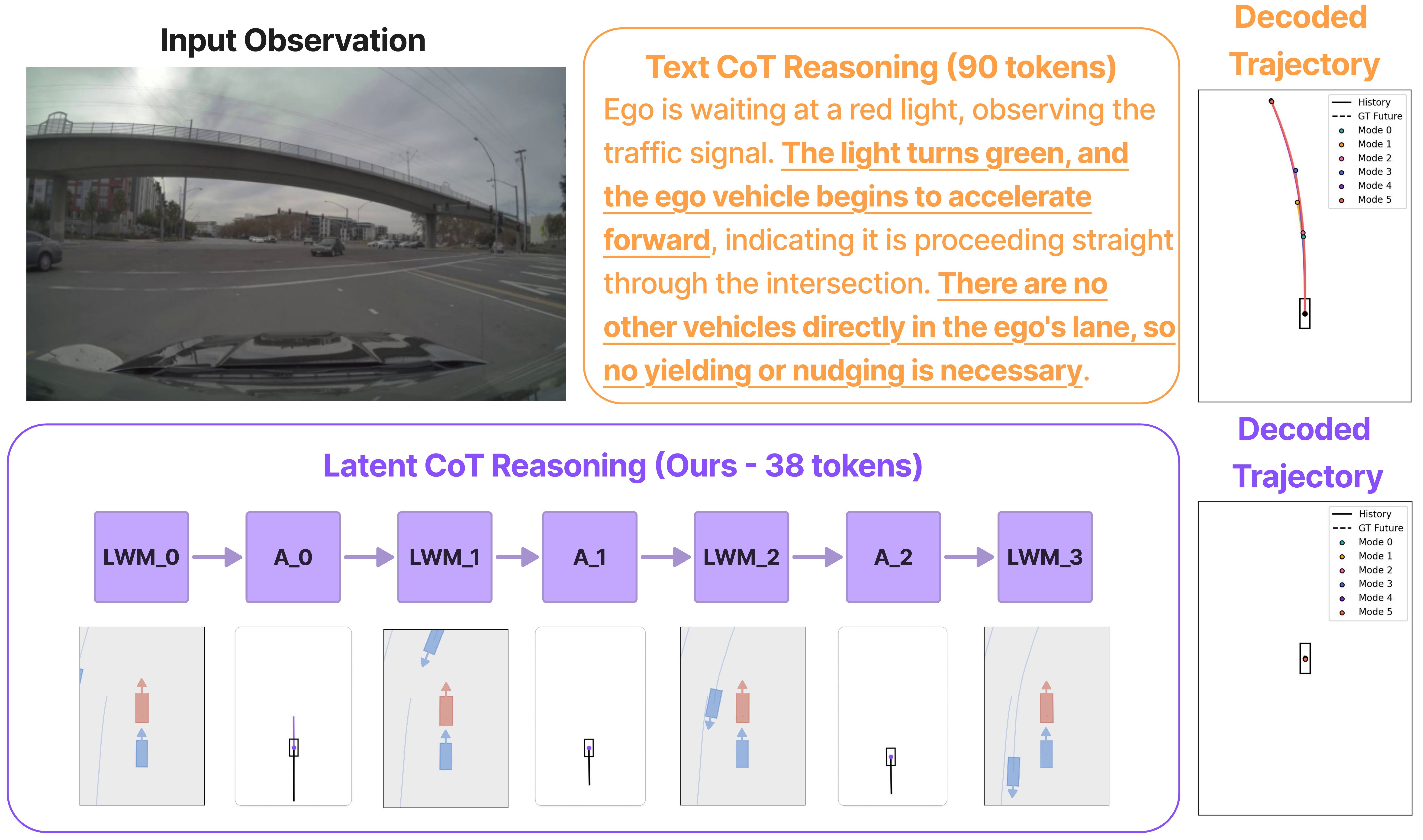}
    \caption{\textbf{Latent Chain-of-Thought Reasoning.}
        Compared to text-based CoT, our proposed Latent CoT provides more efficient and aligned reasoning traces for end-to-end driving VLA models.
        In the above example, the Latent CoT anticipates that the vehicle ahead will cut into the ego lane, indicating that the ego vehicle should hold its position rather than move forward.
        On the other hand, with only Text CoT, the ego vehicle directly drives forward.
    }
    \vspace{-0.2cm}
    \label{fig:teaser}
\end{figure}

In this paper, we propose \METHODNAME, a Latent Chain-of-Thought framework for Driving VLA models. 
Instead of relying on textual CoT, \METHODNAME performs reasoning through vector-space supervised chain-of-thought tokens grounded in a learned \textit{latent world model} (LWM), as shown in~\cref{fig:teaser}.
The latent reasoning process alternates between action-proposal tokens and latent world model prediction tokens, thereby simulating counterfactual futures directly in latent space and using those futures to inform the choice of the next action.
This interleaved latent CoT forms a structured and compact reasoning trace grounded in the multi-agent interaction process, yielding both higher dynamical precision and significantly more efficient inference.
We train \METHODNAME through a three-stage pipeline (\cref{fig:pipeline}).
Starting from a pretrained non-reasoning VLA, we first cold-start with latent CoT by teacher-forcing the model with ground-truth (GT) world model states and reasoning actions proposed by the model itself.
During this process, we simultaneously train a small LWM prediction head to predict LWM embeddings from proposed actions during inference.
Next, we apply reinforcement learning (RL)~\cite{kaelbling1996reinforcement} to refine this initial scaffold of latent reasoning and improve final action prediction using trajectory-level rewards.

We evaluate \METHODNAME on the large-scale PhysicalAI-AV dataset~\cite{nvidia2025avdata}, consisting of 1727 hours of driving data across challenging urban scenarios with dense multi-agent interactions.
In~\cref{tab:main}, we show that \METHODNAME improves trajectory fidelity and driving success compared to the baseline Text CoT VLA models.
Qualitative rollouts in~\cref{fig:quali} show how coherent Latent CoT reasoning benefits driving performance compared with Text CoT reasoning.
We further include results across different scenario categories as well as extensive ablation experiments to show the superior performance of \METHODNAME.

\noindent \textbf{Contributions.} The main contributions of our work are:
\begin{itemize}
\item We rethink the representation of reasoning in VLA models for E2E driving with \METHODNAME, which conducts latent CoT with latent reasoning tokens strongly aligned with driving actions and a latent world model.
\item We introduce a training framework combining latent CoT cold-start, world model training, and closed-loop RL, finding it especially effective for latent reasoning models.
\item We demonstrate consistent empirical gains on a large, diverse E2E driving benchmark: \METHODNAME delivers faster inference, improved driving quality, and larger improvements under interactive RL compared to non-reasoning and text-reasoning baselines.
\end{itemize}

\section{Related Work}

\paragraph{Driving VLA Models}
E2E driving systems learn a direct mapping from raw sensor inputs to trajectories or controls, aiming to reduce handcrafted components and human bias in the traditional perception--prediction--planning pipeline~\cite{hu2023uniad,weng2024paradrive}. Although this has shown effectiveness in common scenarios, classical E2E models struggle in long-tail driving scenarios due to limited world knowledge and weak reasoning structure.
With the rise of foundation models, recent work has explored using pre-trained LLMs and multimodal LLMs as core building blocks for end-to-end driving policies. Early approaches incorporate these models primarily as backbones while still directly predicting actions from multimodal inputs~\cite{xie2025s4, zhou2025opendrivevla,jiang2025irl,fu2025orion}.
More recent methods introduce textual chain-of-thought \cite{wei2022chain} before action prediction, leveraging the common-sense reasoning capabilities of LLM backbones to improve motion planning, particularly in rare or complex scenarios~\cite{tian2024tokenize, wang2024drivecot, hwang2024emma, zhou2025autovla,wang2025alpamayo}.
Different from previous works, our work departs from text-based CoT in driving VLAs and instead performs reasoning directly in a latent representation space. 

\paragraph{Latent World Models}
An alternative to model-free driving policy learning is to leverage latent world models (LWMs)~\cite{ha2018world, schrittwieser2019mastering}. LWMs learn a generalized latent dynamics function that predicts the action-conditioned future evolution of the environment given current observations and planned actions. 
In autonomous driving, LWMs have recently emerged as flexible dynamic models that complement end-to-end policies. Some works jointly learn latent dynamics and the driving policy from expert demonstrations~\cite{hu2022model, wang2024drivedreamer}, enabling the agent to model multi-agent interactions and future outcomes directly in latent space. 
Other efforts leverage trained latent world models to generate additional demonstrations for data augmentation~\cite{popov2024mitigating, mao2025dreamdrive} or to serve as neural simulators for reinforcement learning--based policy training~\cite{huang2023safedreamer, li2024think2drive}.
These approaches highlight the promise of latent dynamics as a way to introduce structure and interaction-awareness into the learning process.

\paragraph{Language-Free Paradigms for Reasoning}
While textual CoT has become a popular strategy for eliciting reasoning in multimodal models, it is not always an ideal medium for tasks that require geometry understanding and dynamics modeling. In addition, textual CoT often contains many non-essential tokens that do not contribute to the underlying reasoning process, inflating token usage and slowing inference without proportional improvements in decision quality~\cite{feng2025efficient, chen2025unveiling}. Recently, a line of work has begun to explore latent reasoning in LLMs, where intermediate computations are performed directly in latent space rather than in natural language. This paradigm enables more compact and informed reasoning~\cite{deng2024explicit, hao2024training, chen2025reasoning}, often with a more cost-effective inference budget. Building on these ideas, subsequent works extend latent reasoning to vision-language models, achieving latent spatial reasoning~\cite{sun2025latent,li2025latent}.
In this work, we adopt this emerging paradigm within driving foundation models and perform reasoning entirely in latent space, demonstrating that latent reasoning is both more effective and substantially more efficient than textual reasoning for autonomous driving.

\section{\METHODNAME: Driving with Latent CoT}

\begin{figure*}[t]
    \centering
    \includegraphics[width=\linewidth]{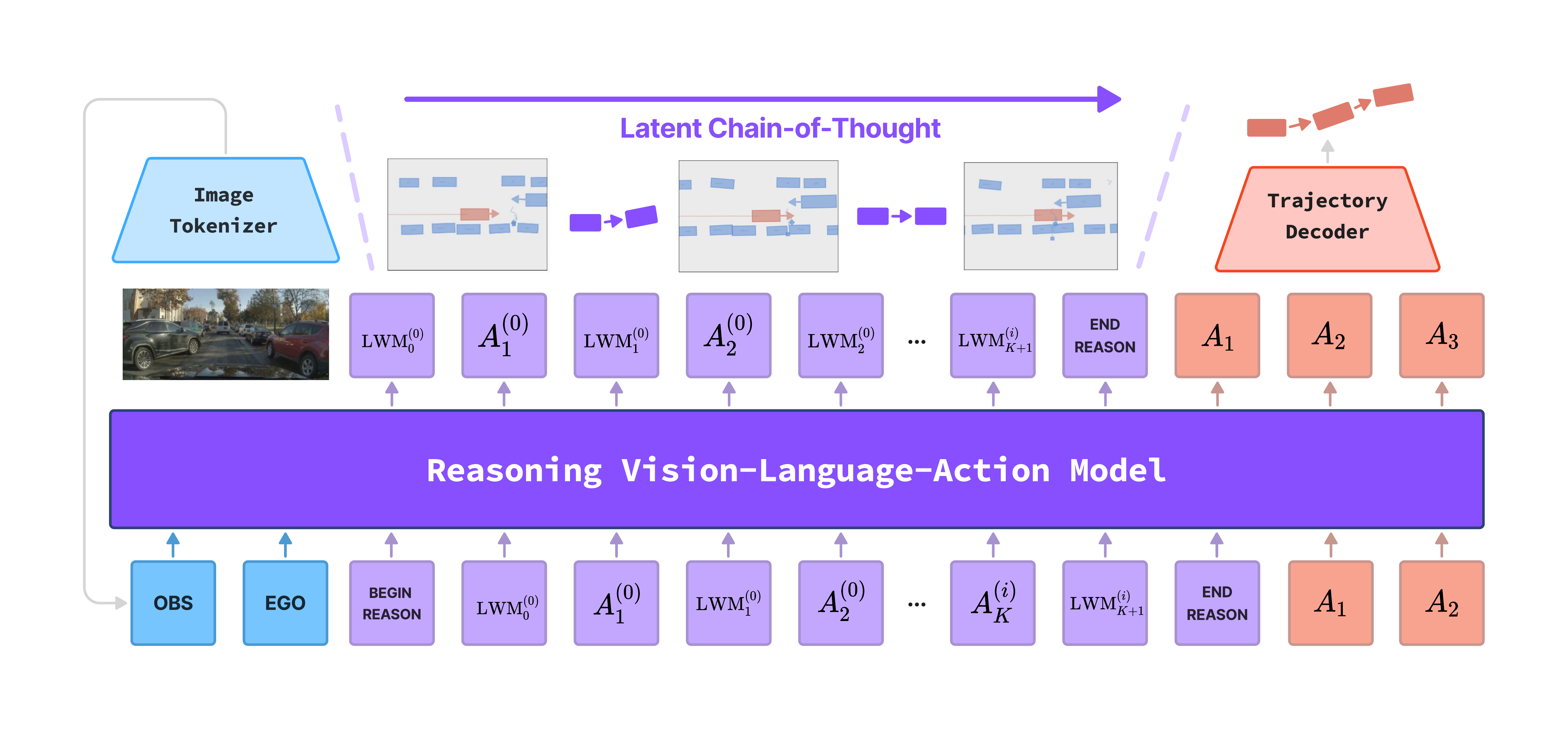}
    \caption{\textbf{Architecture.}
        Overview of our proposed latent reasoning framework.
    }
    \label{fig:architecture}
\end{figure*}

\subsection{Preliminaries}
In this section, we formally define the task, followed by the concepts required to enable latent reasoning.

\paragraph{Task}
We aim to design a policy that maps sensor streams and ego state inputs to future trajectories.
Following previous works on reasoning VLA for driving~\cite{wang2025alpamayo}, we regard E2E driving as modeling an autoregressive distribution over a token sequence that concatenates input information, (optional) reasoning trace, and the future trajectory of the ego vehicle $\tau$:
\begin{equation}
\big[\, o_{\text{image}},\; o_{\text{ego}},\; \textsc{Reason},\; \tau \,\big],
\label{eq:ar1-seq}
\end{equation}
where each component conditions on all previous ones.
The inputs of the model include $o_{\text{image}}$, $M$ front-view (or multi-camera) frames over the last $L$ steps; and $o_{\text{ego}}$, egomotion history.
Given these inputs, the model produces (optional) $\textsc{Reason}$ tokens followed by the future trajectory of the ego vehicle $\tau$. We parameterize $\tau$ as the full 6.4\,\text{s} future at 10\,\text{Hz}, yielding a sequence of 64 future waypoints:
\begin{equation}
\tau=\big\{(x^i,\,y^i,\,\theta^i_{\text{yaw}})\big\}_{i=1}^{64}.
\label{eq:traj}
\end{equation}

\paragraph{Input Tokenizers}
\textit{Image tokenizer:}
Following standard VLM practice, each frame in $o_{\text{image}}$ is tokenized independently using a ViT-based encoder (e.g.,~\cite{bai2025qwen2p5,qwen2025qwen3vl}), producing a sequence of visual tokens
$\;o_{\mathrm{img}}=\mathrm{Tok}_{\mathrm{img}}\!\big(V_{t-L:t}^{1:M}\big)$. Tokens from different camera views and timestamps are concatenated to form the full visual token sequence.
\textit{Egomotion tokenizer:}
The ego vehicle's historical kinematics (speed, yaw rate, past $k$ control actions) are embedded into a compact set of tokens $\;o_{\mathrm{ego}}=\mathrm{Tok}_{\mathrm{ego}}(e_t)$ with learned positional encoding.

\paragraph{Trajectory Tokenizer}
The $6.4$\,s future trajectory at $10$\,Hz is represented using 64 discrete trajectory tokens
$\tau = a_{1:64}$, one token per time step. Each $a_i$ indexes a motion-primitive bin corresponding to the
ego-frame $\Delta$-pose $(\Delta x,\Delta y,\Delta\psi)$.
We build a 1024-code vocabulary via $k$-means on training $\Delta$-poses.
We encode continuous trajectories by quantizing them to indices $a_{1:64}$ with nearest-code assignment.
We decode discrete indices back to $\Delta$-poses via codebook lookup and integrate them over time to recover continuous trajectories $\hat{\tau}$.

\paragraph{Latent World Model (LWM)}
We introduce an ego-centric latent world model state $\mathrm{LWM}_t$ that captures vectorized agent boxes and poses from online perception.
Each $\mathrm{LWM}_t$ summarizes a fixed 1.0\,s window at 10\,Hz (10 frames) as a fixed-size set of vectorized representations of the $K_{\text{agents}}$ nearest non-ego agents.
$\mathrm{LWM}_0$ encodes the most recent history window up to the current time, which \emph{starts} the reasoning process. It can be \emph{given} from online perception (detection, tracking) or \emph{predicted} by the VLA model itself.
$\mathrm{LWM}_1,\mathrm{LWM}_2,\ldots$ represent future $1.0$\,s windows produced during latent reasoning, conditioned on proposal actions.
We encode the tracked agents in each window using an agent encoder initialized from an internal, pretrained trajectory prediction model.
The encoder was trained end-to-end through trajectory-prediction supervision, and we reuse only this agent-encoding branch for LCDrive.
It maps each agent's temporal history to one token, after which a lightweight attention compressor aggregates the agent tokens into a fixed-size set of latent world-model tokens. Architectural details are provided in~\cref{sec:lwm_encoder}.

\paragraph{Reasoning Tokens}
The presence of \textsc{Reason} is optional and used differently across different models.
For the \emph{non-reasoning} baseline model, we set $\textsc{Reason}=\varnothing$.
For a fair comparison, the baseline may \emph{optionally} condition on \emph{only} $\textsc{Reason} = \big[\mathrm{LWM}_0 \big]$ as context.
For \emph{text-based CoT} models (e.g., AR1~\cite{wang2025alpamayo}), \textsc{Reason} consists of a sequence of natural-language tokens that verbally describe intermediate reasoning before action prediction.
In this paper, we propose \emph{latent CoT}, where \textsc{Reason} is instantiated as a short interleaved sequence of latent tokens composed of \emph{action-proposal} tokens and counterfactual latent world-model tokens, initialized from the latent state $\mathrm{LWM}_0$.
By default, $\mathrm{LWM}_0$ is predicted by the VLA model itself given the sensor inputs as context.
We detail the construction of latent \textsc{Reason} tokens in the following section.

\subsection{Latent Chain-of-Thought Reasoning}
We aim to design a compact, action-aligned reasoning process that performs latent counterfactual rollouts in the latent world model state, and keeps the CoT in the same vocabulary as the final trajectory output.

\paragraph{Token Scheme}
We represent each reasoning branch as an interleaved action and latent world model trace \(R^{(i)}\):
\begin{equation}
\label{eq:cot_tokens}
R^{(i)} = \big[ 
A_0^{(i)}, \mathrm{LWM}_1^{(i)}, A_1^{(i)},\mathrm{LWM}_2^{(i)},\ldots,A_{K-1}^{(i)}, \mathrm{LWM}_K^{(i)}\big].
\end{equation}
Here $A_t^{(i)}$ are \emph{action-proposal} tokens drawn from the same action vocabulary as the final output, but grouped as a 1.0s block of 10 stepwise tokens:
\[
A_t^{(i)} \;:=\; \big(a_{10(t-1)+1},\ldots,a_{10t}\big),
\]
which makes proposals easy to produce and interpret.
$\mathrm{LWM}^{(i)}_{t+1}$ is the ego-centric latent world state summarizing the \emph{same} 1.0\,s window at 10\,Hz. 
Reasoning is seeded by the history anchor $\mathrm{LWM}_0$, after which we interleave $(A_t^{(i)},\mathrm{LWM}_{t+1}^{(i)})$ for $t=1\dots K$ to form $R^{(i)}$.

\paragraph{Action Proposal}
At step \(t\), the VLA proposes \(A_t^{(i)}\) conditioned on sensor tokens, the current world state, and the prior reasoning token sequence:
\[
A_t^{(i)} \sim \pi_\theta\!\big(\cdot \,\big|\, o_{\text{image}},\,o_{\text{ego}},\,\mathrm{LWM}_0,\,R^{(i)}_{<t}\big).
\]
Note that $A_t^{(i)}$ uses the same token vocabulary as the final trajectory prediction $\tau$. These proposals are only used as reasoning context and do \emph{not} commit to a specific final plan.

\paragraph{LWM Prediction}
Given the proposal as context, we predict the next latent world state:
\[
\mathrm{LWM}^{(i)}_{t+1} \sim q_\phi\!\Big(\cdot \,\Big|\, o_{\text{image}},\,o_{\text{ego}},\,\mathrm{LWM}_0,\,R^{(i)}_{<t},\, A_t ^{(i)}\Big).
\]
In practice, we compute it with
$f_\phi(\mathbf{h}^{\mathrm{VLA}}_t)$,
where \(\mathbf{h}^{\mathrm{VLA}}_t\) is the VLA hidden embedding after taking \(A_t^{(i)}\) as input and \(f_\phi\) is a lightweight MLP that outputs LWM tokens.

\paragraph{Multi-Branch Reasoning}
To allow the model to spend more reasoning tokens on diverse strategies and paths, we enable autoregressive generation of a fixed number of branches \(B\) (default \(B=2\)).
All branches share the history anchor \(\mathrm{LWM}_0\) and are generated sequentially: for \(i=1\ldots B\), we produce \(R^{(i)}\) while conditioning on previously formed traces \(R^{(<i)}\).
This lets the model refer to prior latent reasoning when proposing the next branch, promoting diversity and yielding more plausible, complementary counterfactual futures under a bounded token budget.
In this paper, we fix both \(K\) and \(B\) at training and evaluation for simplicity.

\paragraph{Action Prediction}
The complete reasoning context is
\[
\textsc{Reason} \;=\; \big[\, \mathrm{LWM}_0,\; R^{(1)},\; \ldots,\; R^{(B)} \,\big].
\]
Conditioned on the sensor input and \(\textsc{Reason}\) in \cref{eq:ar1-seq}, the model predicts the 64 stepwise trajectory tokens \(a_{1:64}\) and decodes the final trajectory \(\hat{\tau}\).
The final actions attend to \emph{all} proposals and their associated latent world model rollouts, forming rich counterfactual context that we will show yields higher-fidelity, safer, and more stable trajectories.

\begin{figure}[t]
    \centering
    \includegraphics[width=\linewidth]{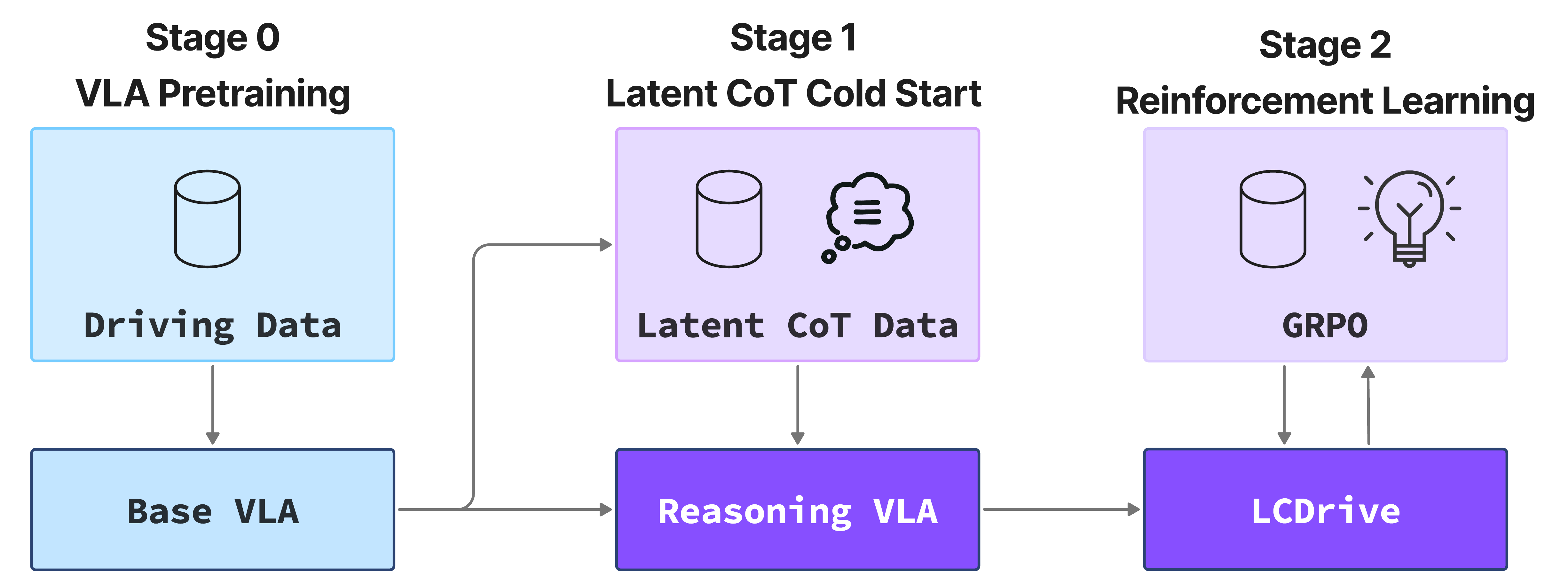}
    \caption{
        \textbf{Training strategy}. 
        We first use a base non-reasoning VLA to create latent CoT data, and cold start \METHODNAME by supervised learning.
        Then, we conduct reinforcement learning to activate useful reasoning capacity of \METHODNAME.
    }
    \label{fig:pipeline}
\end{figure}

\subsection{Training Strategy}
We train \METHODNAME in three training stages (\cref{fig:pipeline}).

\subsubsection{Stage 0 - Non-reasoning Pretraining}
We start from a \emph{non-reasoning} VLA (\textsc{Reason} \(=\varnothing\)) trained via supervised fine-tuning to predict trajectory tokens from driving data.
We keep two copies of this model: (1) one serves as the initialization for \METHODNAME in the later fine-tuning stage; (2) the other is \textit{frozen} and used solely to generate action-proposal tokens for latent reasoning.

\subsubsection{Stage 1 - CoT Cold Start}
In this step, we aim to teach the VLA model the format and structure of latent CoT with teacher forcing.
To this end, we construct supervision data for latent CoT \textsc{Reason} tokens through the following steps.

\paragraph{Action Proposals} 
Given sensor inputs, we use the \emph{frozen} non-reasoning VLA \(\pi_{0}\) to sample \(B\) different trajectories
\(\{\tilde{a}^{(i)}_{1:64}\}_{i=1}^{B}\) in random order.
Each sample is sliced into \(K\) 1.0\,s action blocks:
$
\tilde{A}^{(i)}_{t} := \big(\tilde{a}^{(i)}_{10t+1},\ldots,\tilde{a}^{(i)}_{10t+10}\big).
$

\paragraph{Action-conditioned LWM targets}
For each block \(\tilde{A}^{(i)}_{t}\), we integrate its ego-frame \(\Delta\)-poses to obtain the ego pose for that 1.0\,s window, re-center the \textit{GT} future tracked agent bounding boxes into this ego frame, and encode them to produce a target latent world state:
$\tilde{\mathrm{LWM}}^{(i)}_{t+1}$.
This yields branch-specific world tokens \(\{\tilde{\mathrm{LWM}}^{(i)}_{t+1}\}\) that reflect the \emph{consequences} of each proposal window.

\paragraph{Supervision sequence}
Action proposals and targets are interleaved to form \(B\) reasoning traces \(R^{(i)}\) (\cref{eq:cot_tokens}).
The full training sequence in \cref{eq:ar1-seq} thus becomes
\[
\big[o_{\text{image}},\,o_{\text{ego}},\,
\underbrace{\mathrm{LWM}_0,\; R^{(1)},\ldots,R^{(B)}}_{\textsc{Reason}},\,
a_{1:64}\big].
\]
We input this full sequence to \METHODNAME during training.

\paragraph{Objective}
We train \METHODNAME to minimize a standard  cross-entropy loss over proposals and the final action plan:
\[
\mathcal{L}_{\text{token}}
= \sum_{i=1}^{B}\sum_{t=0}^{K-1}\mathrm{CE}\!\big(A^{(i)}_{t},\tilde{A}^{(i)}_{t}\big)
+ \mathrm{CE}\!\big(a_{1:64},a^{\star}_{1:64}\big).
\]
Additionally, we train the LWM prediction module to predict the corresponding ground-truth LWM embedding during reasoning as well as the initial $\mathrm{LWM}_0$:
\[
\mathcal{L}_{\text{lwm}}
= \|\mathrm{LWM}_0-\tilde{\mathrm{LWM}}_0\|_2^2
+ \sum_{i,t}\|\mathrm{LWM}^{(i)}_{t+1}-\tilde{\mathrm{LWM}}^{(i)}_{t+1}\|_2^2.
\]
The overall objective of \METHODNAME in Stage 1 is:
\begin{equation}
\label{eqn:stage1_loss}
\mathcal{L}_{\text{stage-1}} = \mathcal{L}_{\text{token}} + \lambda \mathcal{L}_{\text{lwm}}.
\end{equation}

\subsubsection{Stage 2 -  Reinforcement Learning}
The second stage post-trains \METHODNAME to actively produce useful latent reasoning and output better actions.
By directly encourage the model to improve the feasibility of the final action \textit{conditioned} the latent reasoning process, the model learns how to produce reasoning tokens beyond imitating the frozen model in Stage 1.

\paragraph{Rollout}
For each training input, we keep the fixed reasoning budget \((K,B)\) and generate a group of \(G\) stochastic \textit{completions}: the policy autoregressively generates \emph{action-proposal} blocks interleaved with latent world states to form branch traces \(R^{(i)}\), and concatenates them into
$
\textsc{Reason}=\big[\mathrm{LWM}_0,R^{(1)},\ldots,R^{(B)}\big].
$
Conditioned on \textsc{Reason} and the sensor tokens, the policy then produces the 64 trajectory tokens \(a_{1:64}\) and decodes \(\hat{\tau}\).

\paragraph{Reward}
We use a single trajectory-accuracy signal: \emph{Average Displacement Error (ADE)} in meters between the predicted and expert trajectories over the 6.4\,s horizon:
\[
\mathrm{ADE}(\hat{\tau},\tau^\star)
=\frac{1}{64}\sum_{i=1}^{64}\left\|\,\hat{\mathbf{p}}_i-\mathbf{p}^{\star}_i\right\|_2,
\]
where $\mathbf{p}_i$ is the $i$-th 2D ego location along the trajectory.
The reward for completion \(j\) is \(R^{(j)}=-\mathrm{ADE}(\hat{\tau}^{(j)},\tau^\star)\).

\paragraph{Learning Algorithm} We use Group Relative Policy Optimization (GRPO)~\citep{shao2024deepseekmath} for RL training. Specifically, for each training example,
we sample a group of \(G\) completions \(\{\hat{\tau}^{(j)}\}_{j=1}^{G}\), compute a trajectory-centric reward \(R^{(j)}\), and construct centered advantages for each completion:
\(A^{(j)} = R^{(j)} - \frac{1}{G}\sum_{k}R^{(k)}\).
We then maximize the advantage-weighted log-probability of the \emph{generated} tokens, including both proposal and final action tokens:
\begin{equation}
\mathcal{L}_{\text{GRPO}}
= - \frac{1}{G}\sum_{j=1}^{G} A^{(j)}
\!\!\sum_{t}
\log \pi_\theta\!\big(x^{(j)}_t \mid \text{context}^{(j)}_t\big).
\end{equation}
Empirically, we found that GRPO performs best without KL regularization, so we omit the KL term in the final objective.
Note that Stage~2 can also be applied to a non-reasoning baseline with \(\textsc{Reason}=\varnothing\).
We will show in \cref{sec:expt_main_results} that RL yields substantially larger gains for \METHODNAME than the baseline.

\label{sec:formatting}

\section{Experiments}

\subsection{Setup}
\paragraph{Dataset}
We conduct our experiments on the recently released PhysicalAI--AV dataset~\cite{nvidia2025avdata}.
It provides large-scale (1700+ hours) real-world multi-camera driving logs with precise ego trajectories and dense multi-agent annotations, enabling realistic end-to-end driving evaluation.
In coordination with the dataset authors, we obtained a \emph{scenario-balanced} subset that maintains consistency with the official public splits of the full dataset: 39{,}072 training clips (87 hours) and 23{,}758 validation clips (53 hours). For each clip, we consider 1.6\,s of history and 6.4\,s of future ego and surrounding-agent trajectories at 10\,Hz.

As summarized in \cref{tab:scenario}, the subset is constructed to balance nominal and eventful scenes: 30\% of clips are \emph{General Driving} and the remaining 70\% are evenly distributed across 14 specific scenarios (e.g., lane keeping, intersection navigation, merges, cut-ins), with 5\% of the data per category.
In addition to its significantly larger scale compared to prior E2E driving validation benchmarks (e.g. nuScenes~\cite{caesar2020nuscenes} with only 150 validation clips, less than 1 hour), this split provides a near-uniform scenario distribution. It avoids dominance by easy cases (e.g., 73.9\% straight driving in nuScenes~\cite{li2024egostatus}) and enables a fair, per-scenario evaluation of driving models.

\paragraph{Metrics}
For each input clip, we randomly sample 6 trajectories from the evaluated model. 
Metrics are then computed for each sample, and the average over all samples is taken to be the overall score of the clip.

To measure the similarity of the model output with the expert driving behaviors, we report ADE (meters) as the mean $\ell_2$ error between the predicted ego positions and expert positions at 10\,Hz over the $T = 64$ steps.
We also measure the safety of the model driving behavior: OffRoad$_{2.5}$ and OffRoad$_{5.0}$ (\%) are the fraction of clips for which \emph{any} point in the predicted ego footprint leaves the drivable area within the first $T\!\in\!\{2.5,5.0\}$ seconds.
Coll@2.5 and Coll@5.0 (\%) are the fraction of clips that experience \emph{any} intersection between the ego polygon and any other agent polygon within the same $S\in\{2.5,5.0\}$\,s window.
Corner Dist (m) measures the mean Euclidean distance between corresponding corners of the predicted and expert ego
boxes (with fixed vehicle dimensions) over the 64 steps at 10\,Hz, capturing both translation and
heading errors. 
More detailed metrics can be found in the supplementary material.

\begin{table*}[t]
\centering
\small
\begin{tabular}{l c c c c c c c c}
\toprule
\textbf{\textsc{Reason}} & \textbf{GT LWM} & \textbf{RL} &
$\mathrm{ADE}$ $\downarrow$ & OffRoad$_{2.5}$ $\downarrow$ & OffRoad$_{5.0}$ $\downarrow$ & Coll$_{2.5}$ $\downarrow$ & Coll$_{5.0}$ $\downarrow$ & Corner Dist. $\downarrow$\\
\midrule
\multirow{2}{*}{LWM$_0$-only$^*$} & \cmark &            & 1.393 & \textbf{1.250} & \textbf{3.104} & \textbf{0.259} & 1.198 & 0.835 \\
                                  & \cmark & \cmark     & 1.397 & 1.430 & 4.218 & 0.326 & 1.706 & 0.948 \\
\midrule
\multirow{2}{*}{Latent CoT$^*$}   & \cmark &            & 1.268 & 1.309 & 3.408 & 0.327 & 0.905 & \textbf{0.691} \\
                                  & \cmark & \cmark     & \textbf{1.197} & 1.303 & 3.443 & 0.318 & \textbf{0.867} & 0.739 \\
\midrule
$\varnothing$ (None)              &        &            & 1.762 & 1.753 & 5.279 & 0.348 & 2.207 & 0.986 \\
Text CoT                          &        &            & 1.650 & 1.391 & \textbf{3.005} & \textbf{0.276} & 0.905 & \textbf{0.642} \\
Latent CoT                        &        &            & 1.668 & 1.268 & 3.536 & 0.322 & 1.591 & 0.904 \\
\textbf{Latent CoT (\METHODNAME)}&        & \cmark     & \textbf{1.626} & \textbf{1.219} & 3.292 & 0.289 & \textbf{0.836} & 0.880 \\
\bottomrule
\end{tabular}
\caption{\textbf{Main evaluation results} on the PhysicalAI-AV dataset~\cite{nvidia2025avdata}. Lower is better for all metrics, bold is best.}
\label{tab:main}
\end{table*}

\begin{table*}[t]
\centering
\small
\begin{tabular}{l ccc|ccc}
\toprule
\multirow{2}{*}{\textbf{Scenario Category}} &
\multicolumn{6}{c}{\textbf{ADE @ 6.4 s (in meters, lower is better)}} \\
\cmidrule(lr){2-7}
& LWM$_0$-only$^*$ & Latent CoT$^*$ & Latent CoT + RL$^*$ & No CoT & Text CoT & \METHODNAME \\
\midrule
General Driving                 & 1.015 & 0.899 & 0.838 & 1.268 & 1.434 & \textbf{1.166} \\
Stop for Vehicle                & 0.760 & 0.542 & 0.514 & 0.995 & \textbf{0.919} & 0.942 \\
Speed Control                   & 1.109 & 1.004 & 1.675 & 1.573 & 2.037 & \textbf{1.376} \\
Nudge Static Obstacle Maneuver  & 1.226 & 1.085 & 1.518 & 1.575 & 1.855 & \textbf{1.387} \\
Traffic Control Compliance      & 1.248 & 1.087 & 0.870 & 1.627 & \textbf{1.312} & 1.431 \\
Vulnerable Road Users (VRU)     & 1.369 & 1.215 & 1.246 & 1.850 & \textbf{1.655} & 1.707 \\
Lead Vehicle Following          & 1.421 & 1.305 & 1.112 & 1.861 & \textbf{1.455} & 1.708 \\
Intersection Navigation         & 1.456 & 1.300 & 1.277 & 1.887 & \textbf{1.725} & 1.730 \\
Lane Keeping                    & 1.535 & 1.484 & 1.420 & 2.021 & \textbf{1.783} & 1.828 \\
Nudge Maneuver                  & 1.541 & 1.453 & 1.554 & 1.966 & 1.909 & \textbf{1.824} \\
Lane Keeping Curve              & 1.675 & 1.553 & 1.857 & 2.162 & 2.002 & \textbf{1.986} \\
Merging                         & 1.716 & 1.571 & 1.375 & 2.215 & 2.169 & \textbf{2.089} \\
Turning Maneuver                & 1.839 & 1.652 & 2.041 & 2.347 & \textbf{1.990} & 2.085 \\
Cut-In                          & 2.076 & 1.913 & 1.220 & 2.583 & \textbf{1.884} & 2.385 \\
Lane Change                     & 2.053 & 1.922 & 1.897 & 2.579 & \textbf{2.167} & 2.423 \\
\midrule
\textbf{Overall}                & 1.397 & 1.268 & 1.197 & 1.762 & 1.650 & \textbf{1.626} \\
\bottomrule
\end{tabular}
\caption{\textbf{ADE split by scenario.} Columns are ordered with methods using GT~LWM (marked with $^*$) shown first. Bold is best.}
\label{tab:scenario}
\end{table*}

\paragraph{Baselines}
All variants share the same non-reasoning backbone, trajectory tokenizer, and decoder. 
Unless noted, training uses the PhysicalAI split mentioned aboave.
All models receive identical inputs and differ only in the format of the \textsc{Reason} tokens.
We compare 
1) \textbf{No CoT} (\(\varnothing\)): VLA without any reasoning tokens;
2) \textbf{LWM\(_0\)-only}: the model conditions on the history latent world model state \(\mathrm{LWM}_0\) but performs no interleaved rollout;
3) \textbf{Latent CoT}: our interleaved action-proposal and latent world-model tokens, initialized from \(\mathrm{LWM}_0\);
4) \textbf{Text CoT}: a language-reasoning baseline that uses English text for reasoning. %(AR1 causal reasoning model). 
We mainly compare methods that \emph{predict} all the LWM tokens needed in the reasoning stage. 
To show performance upper-bounds, we also compare with methods that take \emph{GT} LWM tokens within the reasoning space, marked with $^*$.
Our model, \METHODNAME, is \textbf{Latent CoT} with \emph{Predicted LWM}; we also report performance with and without the RL training stage.

\begin{figure*}[t]
    \centering
    \includegraphics[width=\linewidth]{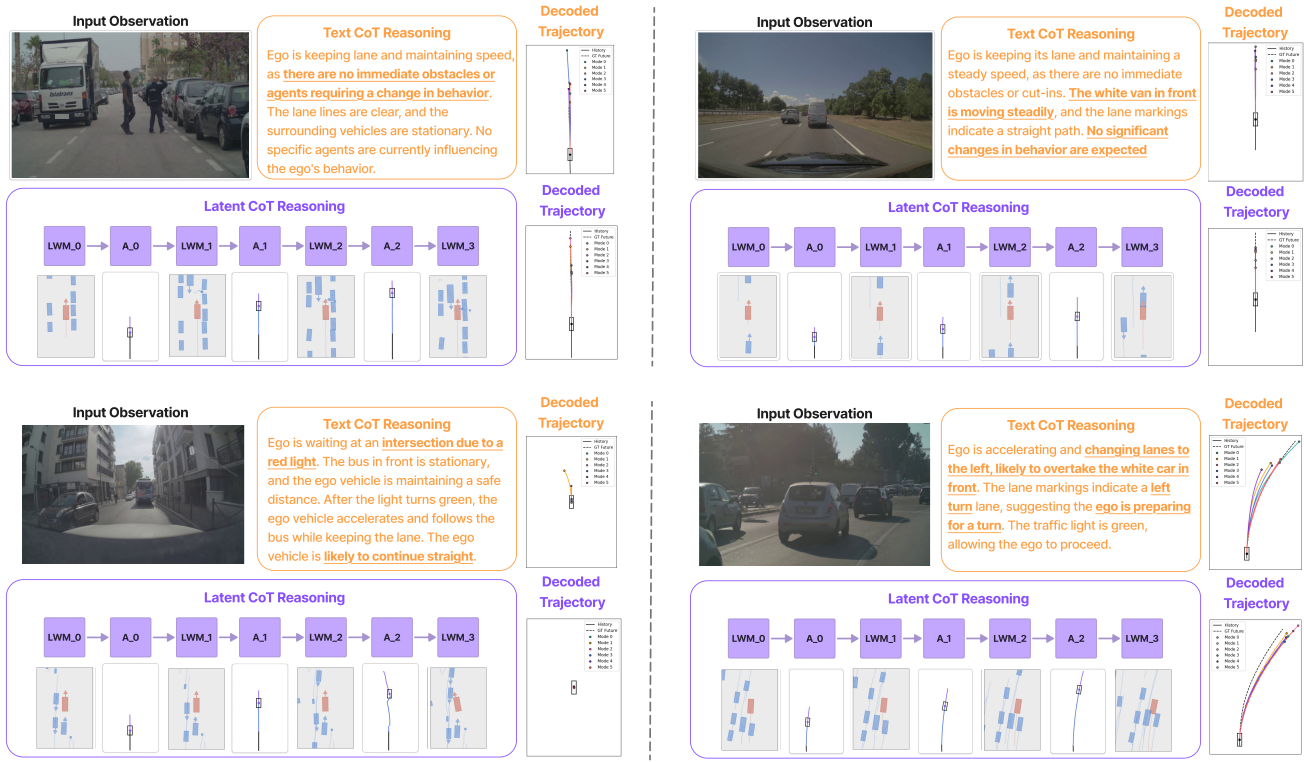}
    \caption{\textbf{Qualitative Results.} Qualitative comparison of textual and latent reasoning in driving VLA models. Latent CoT captures fine-grained spatial relationships and multi-agent interactions while using a smaller inference budget, leading to more stable and accurate trajectory predictions.
    In each case, we highlight the main misalignment of the Text CoT reasoning with the final trajectory. 
    }
    \label{fig:quali}
\end{figure*}

\paragraph{Text CoT baseline} Since obtaining Text-CoT labels for the PhysicalAI-AV dataset~\cite{nvidia2025avdata} is non-trivial, we use model weights provided by the AR1 team~\cite{wang2025alpamayo}.
The model shares the same AR1 architecture, and is pretrained on a large proprietary dataset of driving logs that is an over $100\times$ larger superset of our training set, followed by finetuning on a smaller set of Text-CoT--paired data (though still $\sim10\times$ larger than our training set).
Given its substantially larger training corpus and direct supervision on carefully-curated text CoT dataset, this baseline is expected to perform better than models trained only on PhysicalAI-AV.

\paragraph{Implementation}
We adopt a Qwen3-0.5B~\cite{qwen3} LLM as the language--action module and a DINOv2~\cite{oquab2023dinov2} ViT as the image encoder, following the AR1 architecture design~\cite{wang2025alpamayo}.
Each input clip uses two front-view cameras (wide 120\({}^{\circ }\) and telephoto 30\({}^{\circ }\)) with 320$\times$512 resolution visual inputs. The encoded image tokens are concatenated with ego tokens and \textsc{Reason} tokens before being fed into the decoder.

\textit{Stage-0 non-reasoning pretrain:} We first train a non-reasoning model for 100k steps on the PhysicalAI-AV training split using batch size 128, learning rate 4e-5, and cosine annealing.
\textit{Stage-1 CoT cold start:} We then enable latent reasoning and train for 10k steps with the same optimizer settings.
Action proposals are generated from the frozen non-reasoning model using temperature 0.6 and top-p $= 0.98$.
The loss in \cref{eqn:stage1_loss} is weighted by $\lambda= 0.1$.
\textit{Stage-2 GRPO:} We finally apply RL post-training with GRPO for 3k steps using group size 8, effective batch size 32 sampled completions per update, and a learning rate of 1e-6.
We set the reasoning depth $K=5$ and branch factor $B = 2$ through our experiments unless otherwise specified.

For all approaches, we use temperature 0.6 and top-p $= 0.98$ during sampling of the 6 trajectories per input.

\subsection{Main Results}
\label{sec:expt_main_results}

\paragraph{PhysicalAI-AV evaluation}
We show the main result in Table~\ref{tab:main}.
We first compare the oracle models that use the LWM states (GT LWM).
When provided with ground-truth LWM, Latent CoT$^{*}$ substantially outperforms simply conditioning on the history state (LWM$_0$-only$^{*}$): ADE improves from 1.393 to 1.268, and RL further reduces it to 1.197 while also improving safety (e.g., reducing Coll$_{5.0}$ from 0.905 to 0.867). 
These results indicate that counterfactual reasoning with LWM tokens is an effective substrate for planning with an accurate world state.

Note that RL is beneficial \textit{only} when the model conducts reasoning.
The first two rows show that adding RL to LWM$_0$-only$^{*}$ yields no gain in ADE and worsens OffRoad$_5$, 
whereas RL on Latent CoT$^{*}$ consistently improves both accuracy and safety.
This suggests that RL \emph{activates} a useful latent CoT process and enables closed-loop interactive policy optimization with internal latent rollouts.

In the practical (non-oracle) setting, our model \METHODNAME remains strong. 
\METHODNAME outperforms the non-reasoning baseline by a clear margin (ADE 1.626 vs.\ 1.762; OffRoad$_{2.5}$ 1.219 vs.\ 1.753; Coll$_5$ 0.836 vs.\ 2.207), indicating that learned LWM tokens are highly informative at inference time. 
Notably, the latent CoT process is \emph{robust} to noise in the predicted LWM. Despite errors during model prediction, the interleaved Latent CoT yields consistent gains over the non-reasoning policy. 
Moreover, adding RL on top of predicted LWM further improves accuracy and safety, delivering a clear additional gain. This demonstrates that RL remains beneficial even when the world model is learned, and that it helps the policy exploit the latent CoT interface more effectively.

Compared with the Text CoT baseline, \METHODNAME is comparable without RL and clearly better with RL. 
Before RL, \METHODNAME (ADE 1.668) is on par with Text CoT (1.650). 
After RL, \METHODNAME achieves 1.626 ADE and lower risk (OffRoad$_{2.5}$ 1.219 vs.\ 1.391; Coll$_5$ 0.836 vs.\ 0.905), despite Text CoT being trained on a \textit{much} larger, CoT-annotated dataset. 

Overall, we conclude that (1) LWM tokens provide a more effective reasoning medium than text; (2) RL is especially impactful when paired with latent CoT, reliably translating internal rollouts into better final actions, and (3) introducing latent CoT consistently improves driving quality over its non-reasoning counterpart for driving VLAs.

\paragraph{Scenario breakdown}
We further evaluate \METHODNAME across diverse driving scenarios.
As shown in~\cref{tab:scenario}, \METHODNAME achieves consistent improvements over both non-reasoning and text-reasoning baselines in nearly all categories. 
Compared with the non-reasoning model, \METHODNAME reduces ADE by 7--15\% on most complex maneuvers such as \textit{Intersection Navigation}, \textit{Turning Maneuver}, and \textit{Merging}, which require anticipating multi-agent interactions.
The largest relative gains appear in \textit{Traffic Control Compliance}, \textit{Speed Control}, and \textit{Nudge Static Obstacle Maneuver}, demonstrating the effectiveness of reasoning with LWM which predicts other agents states into the future.

Compared with the Text CoT model, \METHODNAME achieves lower ADE in every scenario, despite Text~CoT being trained on a much larger CoT-annotated corpus. 
Notably, the gaps are largest in interaction-heavy settings such as  \textit{Lead Vehicle Following} (1.708 vs.\ 1.455) and \textit{Stop for Vehicle} (0.942 vs.\ 0.919) indicating that latent reasoning grounded in the LWM space generalizes better to diverse multi-agent behaviors.

The oracle results (Latent CoT$^*$) further illustrate the potential of latent reasoning. 
When supplied with perfect LWM, latent CoT reduces the ADE by large margins across nearly all categories (e.g., 1.300 in \textit{Intersection Navigation} and 0.542 in \textit{Stop for Vehicle}). 
Adding RL on top of oracle LWM yields even stronger results in difficult scenarios such as \textit{Cut-In} (1.220) and \textit{Lane Change} (1.897), demonstrating that latent reasoning becomes especially powerful when accurate multi-agent futures are available.

Overall, the per-scenario analysis shows that latent CoT provides broad, uniform improvements across the full spectrum of driving tasks. 
Reasoning in the latent world-model space leads to better anticipation, more stable long-horizon predictions, and improved performance on categories that require understanding interactions, maneuvers, and compliance with traffic rules. 
These results highlight that latent chain-of-thought is an effective and generalizable reasoning mechanism for VLA-based driving models.

\subsection{Qualitative Results}

In \cref{fig:quali}, we analyze several textual and latent reasoning traces output by the Text CoT baseline and \METHODNAME respectively.
In each example, textual CoT provides a high-level narrative of the environment, but the descriptions remain generic and fail to capture the fine-grained spatial relationships and multi-agent interactions needed for precise driving decisions.
Moreover, these textual rationales often contain numerous non-essential tokens (e.g., stylistic or filler words), which increase inference latency without improving the underlying reasoning.
In contrast, \METHODNAME produces a compact sequence of interleaved action-proposal tokens and latent world-model predictions that encode informative scene dynamics allowing the model to perform multi-step reasoning using only a few compact vector tokens.
Across all examples, \METHODNAME produces motion plan predictions that align closer with the ground truth demonstration while using a significantly lower inference budget.

For each scene, we show one latent world model reasoning trace, selecting the one with the most similar action tokens to the final decoded trajectory.
While \METHODNAME is capable of predicting LWM tokens, it does not require a decoder that reconstructs these tokens back into a human-interpretable visualization.
Therefore, for this comparison, we use the Latent CoT* model from \cref{tab:main} that accesses GT LWM tokens, which we visualize with the corresponding action tokens interleaved.

\section{Conclusion}

In conclusion, we present \METHODNAME: a model that replaces natural language CoT reasoning with a compact, action-aligned latent reasoning space for autonomous driving.
By interleaving action-proposal tokens and world-model tokens, our approach unifies inference-time reasoning and decision making within a single latent world modeling process. This design enables \METHODNAME to reason about the effects of candidate actions via their predicted future outcomes, while avoiding the inefficiencies and potential misalignment of text-based explanations.
Experiments on large-scale real-world driving data demonstrates that latent CoT not only accelerates inference, but also leads to higher-quality trajectories and enables further improvements from closed-loop RL compared to both non-reasoning and text-reasoning baselines.

While these results are encouraging, there are a few limitations that motivate future work: First, training latent CoT currently requires a source of supervision (e.g., GT agent bounding boxes) to ground the representation, which may be difficult to obtain at scale (though recent efforts in autolabeling are addressing this~\cite{sal2024eccv,ravi2024sam2,huang2025vipe,Lee_OpenBox_NeurIPS_2025}). 
Second, our current model does not support easy recovery of a human-interpretable representation from a latent CoT token (e.g., for in-car visualization). Accordingly, building a deeper understanding of the efficiency-interpretability spectrum is an exciting area of future work.
Finally, our model does not yet support flexible reasoning lengths adjusting to different task difficulties, which would make it even more efficient.

\clearpage

\appendix
\renewcommand\thesection{\Alph{section}}
\renewcommand\thesubsection{\Alph{section}.\arabic{subsection}}

\section{Additional Implementation Details}

\subsection{Latent World Model Encoder}
\label{sec:lwm_encoder}
Our latent world model (LWM) encodes the surrounding agents around the ego vehicle (\textit{excluding} the ego vehicle) into a compact set of tokens for latent chain-of-thought reasoning.
Concretely, each LWM state summarizes a fixed $1.0\,\mathrm{s}$ window at $10\,\mathrm{Hz}$ in an ego-centric frame.

\noindent\textbf{Agent encoder pretraining.}
We initialize the per-agent temporal encoder from an internal, pretrained trajectory prediction model.
The pretrained model contains separate encoders for different input modalities, one of which is tracked agent histories.
The input encoders and downstream backbone are optimized end-to-end for trajectory prediction.
Thus, the agent encoder is trained via planning supervision rather than through a separate agent-state or latent-reconstruction objective.
For LCDrive, we reuse only this pretrained agent encoder.

\noindent\textbf{Per-agent temporal encoder.}
For each clip, we select the $N$ nearest agents (based on distance in the current frame).
Each selected agent is encoded independently at this stage; the agent branch performs temporal encoding but does not aggregate information across agents.
The raw per-timestep state of each agent includes position, heading, dimensions, velocity, and other kinematic attributes.
We stack these over a $1.0\,\mathrm{s}$ window (10 frames) to obtain
\[
\texttt{agent\_state} \in \mathbb{R}^{B \times N \times T \times F},
\]
where $B$ is the batch size, $N$ the number of agents, $T{=}10$ the number of timesteps, and $F$ the number of input features.
We first augment the state with $4$ oriented corner points of the 3D bounding box (projected to BEV), resulting in $8$ additional normalized features per timestep.
A linear layer projects the concatenated features from dimension $(F{+}8)$ to a latent dimension $d_{\mathrm{lwm}}$, after which we apply: 1) a learned timestep embedding added along the temporal axis; 2) an agent-type embedding (shared over timesteps) added per agent; 3) a stack of MLP residual blocks along the feature dimension.
This produces a sequence of per-agent, per-timestep features of shape
$\mathbb{R}^{B \times N \times T \times d_{\mathrm{lwm}}}$.

\noindent\textbf{Temporal pooling per agent.}
To summarize the $T{=}10$ timesteps into a single feature per agent, we use a learnable query vector and a cross-attention layer along the time axis.
The query attends to the $T$ timestep features with an attention mask that ignores invalid timesteps, yielding one vector per agent:
\[
\texttt{LWM\_agent} \in \mathbb{R}^{B \times N \times d_{\mathrm{lwm}}}.
\]

\vspace{0.25em}
\noindent\textbf{Two-token LWM summarization.}
The latent world model state $\mathrm{LWM}_t$ used in LCDrive is a compact summary of all agents in the $1.0\,\mathrm{s}$ window.
To introduce cross-agent aggregation, we use a four-layer attention compressor with shared parameters. In each layer, $M \ll N$ learnable query tokens attend to the $N$ agent features through eight-head cross-attention, followed by a feed-forward block. Learned positional embeddings are added to both the agent and query tokens, and pre-normalization is applied.
The resulting representation is
\[
\mathrm{LWM}_t = \texttt{Attn}\bigl(Q_{M},\; \texttt{LWM\_agent}\bigr)
\in \mathbb{R}^{B \times M \times d_{\mathrm{lwm}}}.
\]
These $M$ tokens keep the LWM interface extremely compact for latent reasoning.
In this paper, we use $N=64$ and $M=2$, maintaining a compact representation of LWM while capturing rich agent state information.
The pretrained agent encoder and the newly initialized two-query compressor remain trainable and are jointly optimized during LCDrive Stage~1.

\subsection{Stage 1: CoT Cold Start}

In Stage~1, we teach the model the structure of latent chain-of-thought (CoT) by
\emph{teacher forcing} both the action-proposal tokens and the corresponding latent world model (LWM) tokens.
Here we focus on how we construct the supervised reasoning sequence.

\vspace{0.25em}
\noindent\textbf{Action proposals from a frozen GT-LWM model.}
We start from the LWM$_0$-only model with ground-truth LWM inputs (Row~1 of Tab.~1 in the main paper).
This model is trained without latent reasoning and serves as a strong teacher that produces full 6.4\,s trajectories.
Given sensor inputs $(o_{\text{image}}, o_{\text{ego}})$ and the history latent state $\mathrm{LWM}_0$, the frozen teacher
$\pi_0$ autoregressively samples discrete trajectory tokens
\[
{a}_{1:64} \sim \pi_0(\cdot \mid o_{\text{image}}, o_{\text{ego}}, \mathrm{LWM}_0).
\]
For each training clip, we draw $B$ such trajectories
$\{{a}^{(i)}_{1:64}\}_{i=1}^B$ using top-$p$ sampling (temperature $0.6$, $p=0.98$).
Each sampled trajectory is then sliced into $K$ non-overlapping 1.0\,s action blocks of length 10:
\[
{A}^{(i)}_t := \bigl({a}^{(i)}_{10t+1}, \ldots, {a}^{(i)}_{10(t+1)}\bigr),
\qquad t = 0,\ldots,K{-}1.
\]
These blocks define the \emph{target} action-proposal tokens that our latent CoT policy imitates during cold start.

\noindent\textbf{Action-conditioned LWM supervision.}
For each branch $i$ and block index $t$, we construct an LWM supervision token
${\mathrm{LWM}}^{(i)}_{t+1}$ that encodes the \emph{future world state conditioned on the proposal} ${A}^{(i)}_t$.

Starting from the ground-truth ego pose at the beginning of the window, we integrate the sequence of 10 motion-primitive codes in ${A}^{(i)}_t$ to obtain the ego pose trajectory over that 1.0\,s interval.
At each timestep, we 
1) take the ground-truth bounding boxes of all tracked agents from the PhysicalAI-AV dataset;
2) transform these boxes into the ego-centric frame defined by the integrated ego pose (translation and rotation);
3) feed the resulting agent states into the LWM encoder described in the last subsection.
The encoder yields a compact latent world-model summary for that 1.0\,s window,
which we store as the target token $\mathrm{LWM}^{(i)}_{t+1}$.
Repeating this for all blocks $t = 0,\ldots,K{-}1$ produces an interleaved supervision trace
\[
R^{(i)} =
\bigl[
{A}^{(i)}_0,\,
{\mathrm{LWM}}^{(i)}_1,\,
\ldots,\,
{A}^{(i)}_{K-1},\,
{\mathrm{LWM}}^{(i)}_{K}
\bigr].
\]

\subsection{Stage 2: Reinforcement Learning}
For the reinforcement learning stage of \METHODNAME, we adopt the \texttt{cosmos-rl} framework\footnote{\url{https://github.com/nvidia-cosmos/cosmos-rl}} as our RL backbone.
All RL experiments are conducted on a single 8-GPU node.  
We allocate 6 GPUs as rollout actors, each running an independent sampler replica of \METHODNAME in inference mode; 2 GPUs as learners, jointly performing GRPO optimization and broadcasting updated parameters to all actors.
This partitioning enables high-throughput rollout while keeping optimization stable and fully GPU-resident.

The learning objective is the GRPO loss described in the main paper, but applied to \emph{all} latent CoT tokens.
This allows RL to restructure and refine the latent reasoning process itself, beyond imitation from Stage~1.
Empirically, we observe that latent reasoning benefits significantly more from RL than non-reasoning baselines, highlighting the importance of closed-loop optimization through the latent world-model interface.

\section{Reasoning Action Analysis}
To better understand the behavior of latent chain-of-thought reasoning before/after reinforcement learning,
we analyze the relationship between the \textit{proposal actions} generated during the reasoning stage
and the \textit{final action} output by the policy.
For each validation clip, \METHODNAME generates $B{=}2$ reasoning branches, each producing a 50-step rollout trajectory, decoded from the action proposal tokens $A_t^{(i)}$.
The final decoded trajectory has 64 steps; we truncate it to the first 50 steps for consistent comparison.

\begin{table}[t]
\centering
\small
\begin{tabular}{l cc}
\toprule
\textbf{Metric} & \textbf{No RL} & \textbf{With RL} \\
\midrule
Reasoning Diversity           & 0.412 & 0.353 \\
Reasoning--Action Alignment   & 0.614 & 0.581 \\
Reasoning Quality             & 0.976 & 0.961 \\
Final-Action Quality          & 0.784 & 0.749 \\
\bottomrule
\end{tabular}
\caption{
Reasoning action analysis of \METHODNAME with/without RL training, using GT LWM.
All values are ADE (m).
}
\label{tab:reasoning_action_analysis}
\end{table}

Let $\hat{\tau}_0$ and $\hat{\tau}_1$ denote the two proposal rollouts,
$\hat{\tau}_{\mathrm{final}}$ the final action trajectory (trimmed to 50 steps),
and $\tau^\star$ the ground-truth future trajectory.
We define four metrics as below. All metrics are reported as Average Displacement Error (ADE) in meters.

\begin{enumerate}[leftmargin=*]
    \item \textbf{Reasoning Diversity:}
    \[
    \text{Diversity} = \mathrm{ADE}\bigl(\hat{\tau}_0,\, \hat{\tau}_1\bigr),
    \]
    measures how different the two proposal branches are.

    \item \textbf{Reasoning--Action Alignment:}
    \[
    \text{Alignment} = \min_{k\in\{0,1\}}
    \mathrm{ADE}\bigl(\hat{\tau}_{\mathrm{final}},\, \hat{\tau}_k\bigr),
    \]
    measures how closely the final action aligns with at least one proposal.

    \item \textbf{Reasoning Quality:}
    \[
    \text{Quality} = \frac{1}{2} \sum_{k\in\{0,1\}}
    \mathrm{ADE}\bigl(\hat{\tau}_k,\, \tau^\star\bigr),
    \]
    measures how good the proposals are with respect to the ground-truth trajectory.

    \item \textbf{Final-Action Quality:}
    \[
    \text{Final-Action} = \mathrm{ADE}\bigl(\hat{\tau}_{\mathrm{final}},\, \tau^\star\bigr),
    \]
    the standard ADE of the final action relative to ground truth.
\end{enumerate}

\noindent We evaluate \METHODNAME using GT LWM and compare the result with and without RL, and show the result in \cref{tab:reasoning_action_analysis}.
We summarize two key aspects of the reasoning behavior:
(i) how latent reasoning behaves in general, and
(ii) how reinforcement learning further improves it.
Together, these results reveal the functional role of latent chain-of-thought reasoning in \METHODNAME.
We have the following observations:

\textbf{1) Final actions improve upon the reasoning proposals.}
In both settings, we observe that
\text{Final-Action Quality} $<$ \text{Reasoning Quality}.
This means that even though the reasoning branches provide two candidate future plans, the decoder does not simply copy a branch. 
Instead, it selects the more promising proposal and further \emph{refines} it to produce a more accurate final trajectory.
This refinement effect becomes even stronger after RL.

\textbf{2) Strong alignment between reasoning proposals and the final action.}
Across both models, the Reasoning--Action Alignment score remains small,
indicating that the final trajectory lies close to at least one of the proposal branches.
This shows that the proposal actions are actively used.
After RL, the alignment improves (0.614 $\rightarrow$ 0.581),
indicating that RL strengthens the integration between proposals and the final action.
Note that the Reasoning-Action Alignment score is consistently lower than the Reasoning Quality score. 
This means that the final action lies \emph{closer to one of the reasoning proposals} than either proposal lies to the ground truth. 
Thus, the final plan is strongly aligned with the latent reasoning process, showing that \METHODNAME relies on and refines the reasoning rollouts when producing its final trajectory.

\textbf{3) Reasoning branches maintain meaningful diversity.}
The Diversity score for both models indicates the two branches represent distinct motion hypotheses.
This is essential in multi-agent driving scenarios with inherent uncertainty.
RL slightly reduces diversity (0.412 $\rightarrow$ 0.353),
but the branches remain significantly different.
In other words, RL  makes exploration more targeted towards better proposal quality (0.976 $\rightarrow$ 0.961).

Overall, we find that the final action trajectory is tightly aligned with the latent reasoning proposals, yet still achieves clearly lower ADE to the ground truth than the proposals themselves, showing that the model both uses and refines the proposed futures. 
Compared to the latent CoT model without RL, closed-loop RL further reduces both proposal and final-action errors and strengthens the alignment between proposals and the final decision.

\section{Inference Efficiency Study}

\begin{subsection}{Ablation Study on Reasoning Depth}
\begin{figure}[t]
    \centering
    \includegraphics[width=\linewidth]{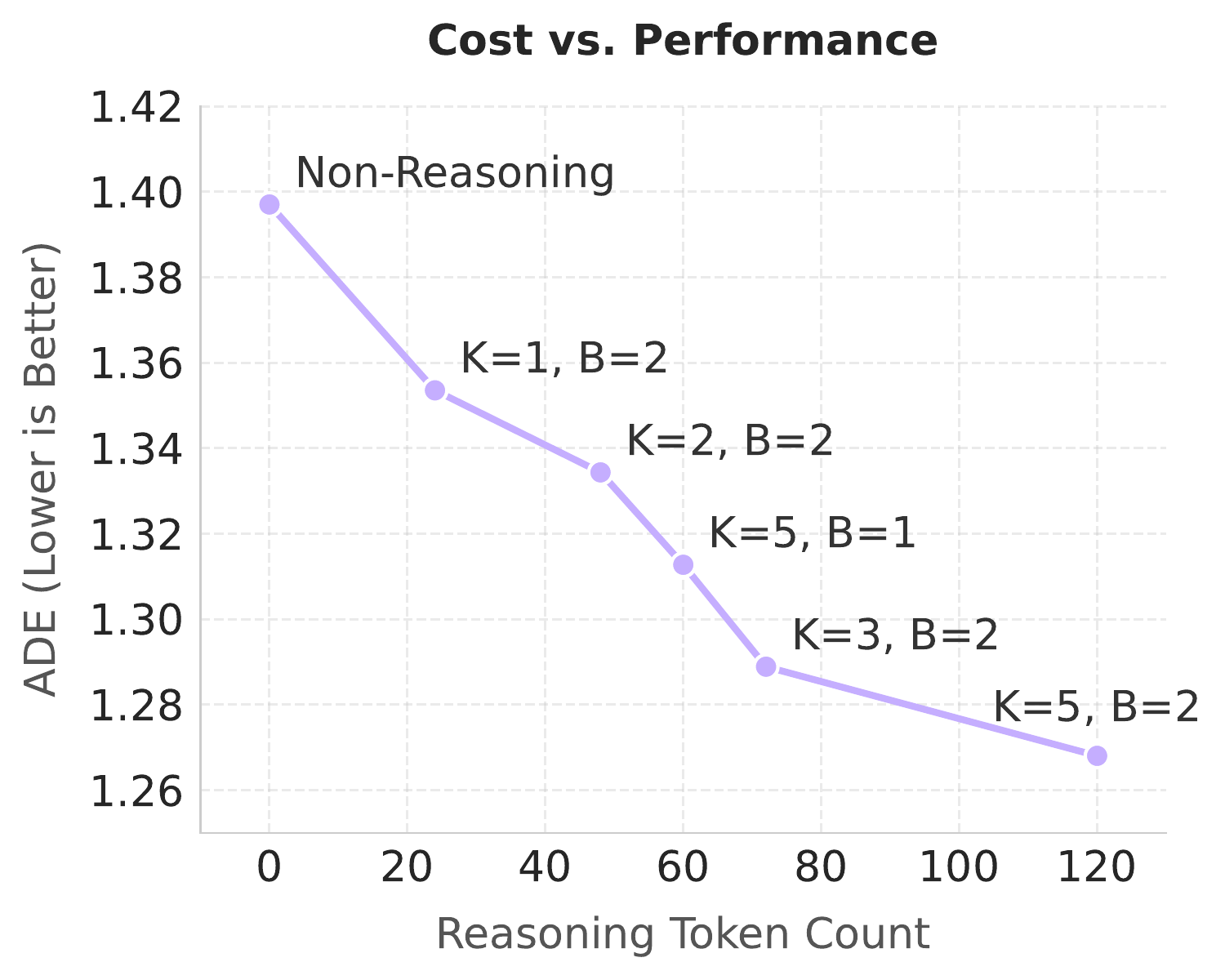}
    \caption{
        \textbf{Efficiency Curve}. 
        We train different variants of \METHODNAME with different reasoning depth $K$ and branch factor $B$.
    }
    \label{fig:efficiency_curve}
\end{figure}

In this section, we study the trade-off between the
reasoning token budget and trajectory accuracy by varying the
reasoning depth $K$ and branch factor $B$ of \METHODNAME (GT LWM, Non-RL).
For each variant, we construct the CoT supervision target in Stage 1 CoT Cold start stage with different settings of $K$ and $B$.
Then, we train the model with teacher forcing with different reasoning depths and branch factors, keeping all other components and hyperparameters fixed across runs.
Importantly, we \textit{do not} apply RL fine-tuning and we \textit{do not}
use predicted LWM tokens in this study, since our goal here is to quantify the tradeoff of reasoning cost and final action performance of latent CoT.

We then evaluate each model on the validation dataset and compare the performance in \cref{fig:efficiency_curve}. We also compare them with the non-reasoning baseline (LWM$_0$-only with GT LWM).
The horizontal axis plots the number of reasoning tokens
generated per input clip, and the vertical axis shows the resulting ADE
(lower is better). 
We have the following observations:

\textbf{1) Latent CoT provides consistent improvements over the baseline}
The leftmost point corresponds to the non-reasoning model. 
Introducing even a minimal amount of
latent reasoning (e.g., $K{=}1$, $B{=}2$ with 24 tokens) produces a clear reduction in
ADE. This demonstrates that a small number of interleaved
action-proposal and latent world-model tokens already provides
useful counterfactual context for the final trajectory prediction.

\textbf{2) Increasing reasoning budget yields meaningful gains}
As we increase $(K,B)$, performance improves smoothly,
indicating that deeper latent reasoning enables the model to explore
more steps into the future and produce better action plans based on that.
The largest gains are obtained when moving from shallow reasoning
(e.g., $K{=}1,2$) to larger reasoning depth ($K{=}3$--$5$).
Beyond this range, improvements are smaller but still positive,
showing that \METHODNAME remains effective with different levels of token budgets.

\textbf{3) Branching ($B$) leads to complementary improvements to depth ($K$)}
Branches encourage diverse counterfactual futures.
Models with multiple branches (e.g., $K{=}5,B{=}2$)
outperform the one with the same depth but fewer branches
(e.g., $K{=}5,B{=}1$).
This aligns with our diversity analysis: exploring alternative
counterfactual futures provides richer reasoning signals for the final policy.

Overall, this curve indicate that latent reasoning offers a highly effective cost-performance tradeoff: a modest reasoning budget (120 tokens) achieves strong trajectory accuracy while remaining relatively cheap.
These results demonstrate that \METHODNAME can flexibly trade inference cost for planning quality.
Even lightweight latent CoT substantially enhances the end-to-end driving performance.

\subsection{Inference Cost Analysis}
We next compare the inference cost of latent chain-of-thought (Latent CoT)
reasoning in \METHODNAME with a text-based CoT baseline.

\paragraph{Latent CoT inference cost}
In \METHODNAME, each reasoning step $k \in \{1,\dots,K\}$ simulates a
$1.0\,\mathrm{s}$ future window and produces:
(i) $10$ discrete action tokens (representing the ego trajectory at 10\,Hz),
and (ii) $2$ latent world model (LWM) tokens.
For a model with reasoning depth $K$ and branch factor $B$, the total number
of latent reasoning tokens is therefore
\[
N_{\text{latent}} \approx (10 + 2) \times K \times B,
\]
plus a small constant overhead for the special tokens.
At inference time, the inference cost of latent reasoning scales linearly with
$N_{\text{latent}}$.

\paragraph{Text CoT baseline cost}
For comparison, we tokenize the text cot reasoning produced by text-CoT
baseline and compute the statistics over the validation dataset.
Over this dataset we obtain an average length of $71.8$ tokens,
a 75-th percentile of $80$ tokens,
and a long tail up to $252$ tokens per clip.
Thus, a typical text-CoT explanation requires on the order of
$70$--$80$ additional tokens at inference time.

From the cost--performance curve in Fig.~\ref{fig:efficiency_curve},
we find that \METHODNAME already achieves \emph{significant}
improvements over the non-reasoning baseline using only a small,
fixed latent budget of roughly $20$--$60$ tokens
(e.g., shallow configurations such as $(K,B)=(1,2)$, $(2,2)$, or $(3,2)$).
These settings use comparable or fewer tokens than typical text CoT, showing that compact latent reasoning is very cost-effective.
As we increase the latent reasoning depth and branch factor, the model
consistently achieves better trajectory accuracy, and remains
\emph{superior} to the text-CoT baseline (as shown in Table 1 of our paper) when using
similar total tokens.
This suggests that latent world-model rollouts provide more actionable
planning signal per token than free-form natural language reasoning.

\paragraph{Potential for further latent reasoning}
Our current action tokenizer produces $10$ tokens per second of motion.
A promising next step is to design a more aggressive motion tokenizer
(e.g., fewer tokens per second or multi-step primitives),
which would \emph{linearly} reduce the latent reasoning token count
for a fixed $(K,B)$.
Because these tokens are structured and low-entropy compared to text,
they are much easier to compress than natural-language CoT,
indicating significant room for future latency and cost reductions
while preserving the benefits of latent reasoning.

\end{subsection}

\end{document}